\pdfoutput=1
\documentclass[11pt]{article}

\usepackage[]{naacl2021}

\usepackage{times}
\usepackage{latexsym}

\usepackage[T1]{fontenc}
\usepackage[utf8]{inputenc}

\usepackage{microtype}
\usepackage{multirow}
\usepackage{booktabs}
\usepackage{graphicx}

\usepackage{latexsym}
\usepackage{mathrsfs}
\usepackage{amsmath}
\usepackage{amssymb}
\usepackage{makecell}
\usepackage{hyperref}

\usepackage{caption}
 \usepackage{subcaption}

\title{Are Multilingual Models Effective in Code-Switching?}

\author{Genta Indra Winata, Samuel Cahyawijaya, Zihan Liu, Zhaojiang Lin, \\\textbf{Andrea Madotto}, \textbf{Pascale Fung} \\
  Center for Artificial Intelligence Research (CAiRE)\\
  The Hong Kong University of Science and Technology\\
  \texttt{giwinata@connect.ust.hk}}

\begin{document}
\maketitle
\begin{abstract}
Multilingual language models have shown decent performance in multilingual and cross-lingual natural language understanding tasks. However, the power of these multilingual models in code-switching tasks has not been fully explored. In this paper, we study the effectiveness of multilingual language models to understand their capability and adaptability to the mixed-language setting by considering the inference speed, performance, and number of parameters to measure their practicality. We conduct experiments in three language pairs on named entity recognition and part-of-speech tagging and compare them with existing methods, such as using bilingual embeddings and multilingual meta-embeddings. Our findings suggest that pre-trained multilingual models do not necessarily guarantee high-quality representations on code-switching, while using meta-embeddings achieves similar results with significantly fewer parameters.
\end{abstract}

\section{Introduction}
Learning representation for code-switching has become a crucial area of research to support a greater variety of language speakers in natural language processing (NLP) applications, such as dialogue system and natural language understanding (NLU). Code-switching is a phenomenon in which a person speaks more than one language in a conversation, and its usage is prevalent in multilingual communities. Yet, despite the enormous number of studies in multilingual NLP, only very few focus on code-switching. Recently, contextualized language models, such as mBERT~\cite{devlin2019bert} and XLM-R~\cite{conneau2020unsupervised} have achieved state-of-the-art results on monolingual and cross-lingual tasks in NLU benchmarks~\cite{wang2018glue,hu2020xtreme,wilie2020indonlu,liu2020attention,lin2020xpersona}. However, the effectiveness of these multilingual language models on code-switching tasks remains unknown.

Several approaches have been explored in code-switching representation learning in NLU. Character-level representations have been utilized to address the out-of-vocabulary issue in code-switched text~\cite{winata2018bilingual,wang2018code}, while external handcrafted resources such as gazetteers list are usually used to mitigate the low-resource issue in code-switching~\cite{aguilar2017multi,trivedi2018iit}; however, this approach is very limited because it relies on the size of the dictionary and it is language-dependent. In another line of research, meta-embeddings have been used in code-switching by combining multiple word embeddings from different languages~\cite{winata2019learning,winata2019hierarchical}. This method shows the effectiveness of mixing word representations in closely related languages to form language-agnostic representations, and is considered very effective in Spanish-English code-switched named entity recognition tasks, and significantly outperforming mBERT~\cite{khanuja2020gluecos} with fewer parameters. 

While more advanced multilingual language models~\cite{conneau2020unsupervised} than multilingual BERT~\cite{devlin2019bert} have been proposed, their effectiveness is still unknown in code-switching tasks. Thus, we investigate their effectiveness in the code-switching domain and compare them with the existing works. Here, we would like to answer the following research question, \textit{``Which models are effective in representing code-switching text, and why?."}

% An extension of the work is proposed by~\citet{winata2019hierarchical} by mixing word, subword, and character representations as hierarchical meta-embeddings. This method applies compositionality into account to provide rich information to the code-switching model by utilizing languages' commonalities. 

\begin{figure*}[t!]
    \centering
    \resizebox{0.99\textwidth}{!}{
        \includegraphics{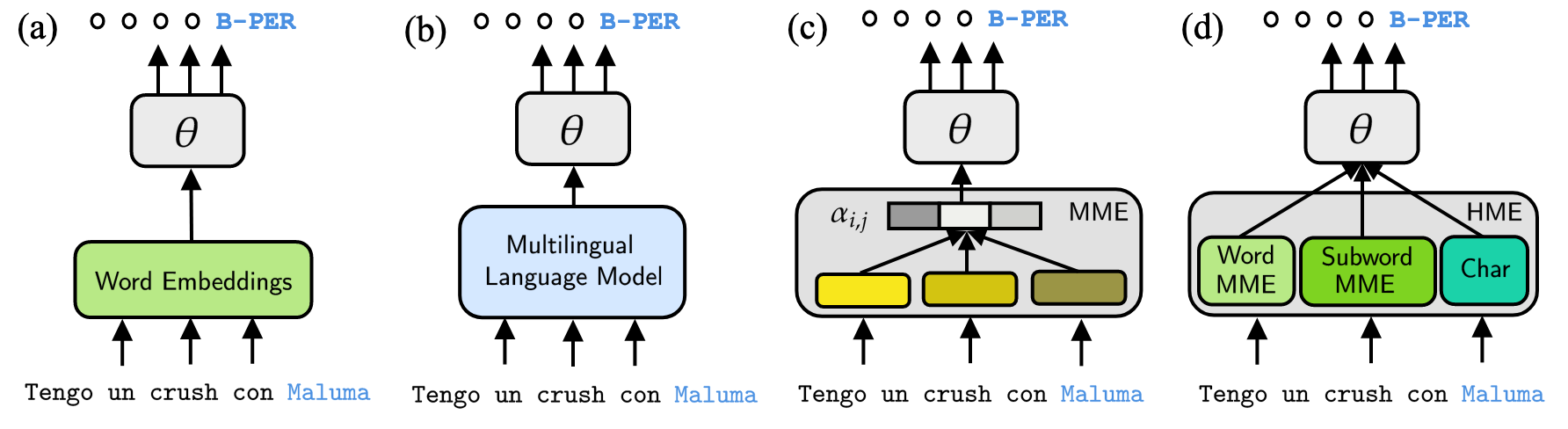}
    }
    \caption{Model architectures for code-switching modeling: (a) model using word embeddings, (b) model using multilingual language model, (c) model using multilingual meta-embeddings (MME), and (d) model using hierarchical meta-embeddings (HME).}
    \label{architecture}
\end{figure*}

In this paper, we evaluate the representation quality of monolingual and bilingual word embeddings, multilingual meta-embeddings, and multilingual language models on five downstream tasks on named entity recognition (NER) and part-of-speech tagging (POS) in Hindi-English, Spanish-English, and Modern Standard Arabic-Egyptian. We study the effectiveness of each model by considering three criteria: performance, speed, and the number of parameters that are essential for practical applications. Here, we set up the experimental setting to be as language-agnostic as possible; thus, it does not include any handcrafted features.

Our findings suggest that multilingual pre-trained language models, such as $\text{XLM-R}_{\text{BASE}}$, achieves similar or sometimes better results than the hierarchical meta-embeddings (HME)~\cite{winata2019hierarchical} model on code-switching. 
% ~\citet{aguilar2020char2subword} showed a possibility to improve the performance of mBERT by introducing continue pre-training and character-level alternation for spelling robustness slightly improves the performance.
On the other hand, the meta-embeddings use word and subword pre-trained embeddings that are trained using significantly less data than mBERT and $\text{XLM-R}_{\text{BASE}}$ and can achieve on par performance to theirs. Thus, we conjecture that the masked language model is not be the best training objective for representing code-switching text. Interestingly, we found that $\text{XLM-R}_{\text{LARGE}}$ can improve the performance by a great margin, but with a substantial cost in the training and inference time, with 13x more parameters than HME-Ensemble for only around a 2\% improvement. 
% approaches (cite). This method is considered to be very effective in many code-switched named entity recognition tasks. It dynamically learns how to mix language representations to form language-agnostic representations. The other direction is to build word embeddings from synthetic code-switching text (cite). 
% In this paper, we provide a thorough analysis on 
The main contributions of our work are as follows:
\begin{itemize}
    \item We evaluate the performance of word embeddings, multilingual language models, and multilingual meta-embeddings on code-switched NLU tasks in three language pairs, Hindi-English (HIN-ENG), Spanish-English (SPA-ENG), and Modern Standard Arabic-Egyptian (MSA-EA), to measure their ability in representing code-switching text.
    \item We present a comprehensive study on the effectiveness of multilingual models on a variety of code-switched NLU tasks to analyze the practicality of each model in terms of performance, speed, and number of parameters.
    \item We further analyze the memory footprint required by each model over different sequence lengths in a GPU. Thus, we are able to understand which model to choose in a practical scenario.
\end{itemize}

\section{Representation Models}
In this section, we describe multilingual models that we explore in the context of code-switching. Figure~\ref{architecture} shows the architectures for a word embeddings model, a multilingual language model, and the multilingual meta-embeddings (MME), and HME models.

\subsection{Word Embeddings}
\subsubsection{FastText}
In general, code-switching text contains a primary language the matrix language (ML)) as well as a secondary language (the embedded language (EL)). To represent code-switching text, a straightforward idea is to train the model with the word embeddings of the ML and EL from FastText~\cite{grave2018learning}. 
Code-switching text has many noisy tokens and sometimes mixed words in the ML and EL that produce a ``new word'', which leads to a high number of out-of-vocabulary (OOV) tokens. To solve this issue, we utilize subword-level embeddings from FastText~\cite{grave2018learning} to generate the representations for these OOV tokens. We conduct experiments on two variants of applying the word embeddings to the code-switching tasks: FastText (ML) and FastText (EL), which utilize the word embeddings of ML and EL, respectively.

\subsubsection{MUSE}
To leverage the information from the embeddings of both the ML and EL, we utilize MUSE~\cite{lample2018word} to align the embeddings space of the ML and EL so that we can inject the information of the EL embeddings into the ML embeddings, and vice versa. We perform alignment in two directions: (1) We align the ML embeddings to the vector space of the EL embeddings (denoted as MUSE (ML $\rightarrow$ EL)); (2) We conduct the alignment in the opposite direction, which aligns the EL embeddings to the vector space of the ML embeddings (denoted as MUSE (EL $\rightarrow$ ML)). After the embeddings alignment, we train the model with the aligned embeddings for the code-switching tasks.

\subsection{Multilingual Pre-trained Models}
Pre-trained on large-scale corpora across numerous languages, multilingual language models~\cite{devlin2019bert,conneau2020unsupervised} possess the ability to produce aligned multilingual representations for semantically similar words and sentences, which brings them advantages to cope with code-mixed multilingual text. 

\subsubsection{Multilingual BERT}
Multilingual BERT (mBERT)~\cite{devlin2019bert}, a multilingual version of the BERT model, is pre-trained on Wikipedia text across 104 languages with a model size of 110M parameters. It has been shown to possess a surprising multilingual ability and to outperform existing strong models on multiple zero-shot cross-lingual tasks~\cite{pires2019multilingual,wu2019beto}. Given its strengths in handling multilingual text, we leverage it for code-switching tasks.

\subsubsection{XLM-RoBERTa}
XLM-RoBERTa (XLM-R)~\cite{conneau2020unsupervised} is a multilingual language model that is pre-trained on 100 languages using more than two terabytes of filtered CommonCrawl data. Thanks to the large-scale training corpora and enormous model size (XLM-R$_\text{BASE}$ and XLM-R$_\text{LARGE}$ have 270M and 550M parameters, respectively), XLM-R is shown to have a better multilingual ability than mBERT, and it can significantly outperform mBERT on a variety of cross-lingual benchmarks. Therefore, we also investigate the effectiveness of XLM-R for code-switching tasks.

\subsubsection{Char2Subword}
Char2Subword introduces a character-to-subword module to handle rare and unseen spellings by training an embedding lookup table~\cite{aguilar2020char2subword}. This approach leverages transfer learning from an existing pre-trained language model, such as mBERT, and resumes the pre-training of the upper layers of the model. The method aims to increase the robustness of the model to various typography styles. 
% \subsubsection{Code-switching ELMo}

\subsection{Multilingual Meta-Embeddings}
The MME model~\cite{winata2019learning} is formed by combining multiple word embeddings from different languages. Let's define $\mathbf{w}$ to be a sequence of words with $n$ elements, where $\mathbf{w} = [w_1,\dots,w_n]$. First, a list of word-level embedding layers is used $E^{(w)}_i$ to map words $\mathbf{w}$ into embeddings $\mathbf{x}_i$. Then, the embeddings are combined using one out of the following three methods: concat, linear, and self-attention. We briefly discuss each method below. 

\paragraph{Concat}
This method concatenates word embeddings by merging the dimensions of word representations into higher-dimensional embeddings. This is one of the simplest methods to join all embeddings without losing information, but it requires a larger activation memory than the linear method.
\begin{equation}
\mathbf{x}_i^{\text{CONCAT}} = [\mathbf{x}_{i,1},...,\mathbf{x}_{i,n}].
\end{equation}

\paragraph{Linear}
This method sums all word embeddings into single word embeddings with equal weight without considering each embedding's importance. The method may cause a loss of information and may generate noisy representations. Also, though it is very efficient, it requires an additional layer to project all embeddings into a single-dimensional space if one embedding is larger than another.
\begin{align*}
\mathbf{x'}_{i,j} &= \mathbf{W}_j \cdot \mathbf{x}_{i,j}, \\
\mathbf{x}_i^{\text{LINEAR}} &= \sum_{j=0}^n{\mathbf{x'}_{i,j}}.
\end{align*}

\paragraph{Self-Attention}
This method generates a meta-representation by taking the vector representation from multiple monolingual pre-trained embeddings in different subunits, such as word and subword. It applies a projection matrix $\mathbf{W}_j$ to transform the dimensions from the original space $\mathbf{x}_{i,j} \in \mathbb{R}^d$ to a new shared space $\mathbf{x'}_{i,j} \in \mathbb{R}^{d'}$. Then, it calculates attention weights $\alpha_{i,j} \in \mathbb{R}^{d'}$ with a non-linear scoring function $\phi$ (e.g., tanh) to take important information from each individual embedding $\mathbf{x'}_{i,j}$. Then, MME is calculated by taking the weighted sum of the projected embeddings $\mathbf{x'}_{i,j}$:
\begin{align}
\mathbf{x'}_{i,j} &= \mathbf{W}_j \cdot \mathbf{x}_{i,j}, \\
\alpha_{i,j} &= \frac{\exp(\phi(\mathbf{x'}_{i,j}))}{\sum_{k=1}^n  \exp(\phi(\mathbf{x'}_{i,k}))}, \\
\mathbf{u}_i &= \sum_{j=1}^n {\alpha_{i,j} 
\mathbf{x'}_{i,j}}.
\end{align}

\subsection{Hierarchical Meta-Embedings}
The HME method combines word, subword, and character representations to create a mixture of embeddings~\cite{winata2019hierarchical}. It generates multilingual meta-embeddings of words and subwords, and then, concatenates them with character-level embeddings to generate final word representations. HME combines the word-level, subword-level, and character-level representations by concatenation, and randomly initializes the character embeddings. During the training, the character embeddings are trainable, while all subword and word embeddings remain fixed.

\subsection{HME-Ensemble}
The ensemble is a technique to improve the model's robustness from multiple predictions. In this case, we train the HME model multiple times and take the prediction of each model. Then, we compute the final prediction by majority voting to achieve a consensus. This method has shown to be very effective in improving the robustness of an unseen test set. Interestingly, this method is very simple to implement and can be easily spawned in multiple machines, as in parallel processes.

\section{Experiments}

In this section, we describe the details of the datasets we use and how the models are trained.

\subsection{Datasets}
We evaluate our models on five downstream tasks in the LinCE Benchmark~\cite{aguilar2020lince}. We choose three named entity recognition (NER) tasks, Hindi-English (HIN-ENG)~\cite{singh2018language}, Spanish-English (SPA-ENG)~\cite{aguilar2018named} and Modern Standard Arabic (MSA-EA)~\cite{aguilar2018named}, and two part-of-speech (POS) tagging tasks, Hindi-English (HIN-ENG)~\cite{singh2018twitter} and Spanish-English (SPA-ENG)~\cite{soto2017crowdsourcing}. We apply Roman-to-Devanagari transliteration on the Hindi-English datasets since the multilingual models are trained with data using that form. Table~\ref{dataset} shows the number of tokens of each language for each dataset. We classify the language with more tokens as the ML and the other as the EL. We replace user hashtags and mentions with \texttt{<USR>}, emoji with \texttt{<EMOJI>}, and URL with \texttt{<URL>} for models that use word-embeddings, similar to~\citet{winata2019learning}. We evaluate our model with the micro F1 score for NER and accuracy for POS tagging, following~\citet{aguilar2020lince}.

\begin{table}[!ht]
\centering
\resizebox{0.45\textwidth}{!}{
\begin{tabular}{lrrll}
\toprule
 & \multicolumn{1}{c}{\textbf{\#L1}} & \multicolumn{1}{c}{\textbf{\#L2}} & \textbf{ML} & \textbf{EL} \\ \midrule
\multicolumn{5}{l}{NER} \\ \midrule
HIN-ENG & 13,860 & 11,391 & HIN & ENG \\
SPA-ENG & 163,824 & 402,923 & \multicolumn{1}{l}{ENG} & \multicolumn{1}{l}{SPA} \\
MSA-EA$^{\dagger}$ & \multicolumn{1}{c}{-} & \multicolumn{1}{c}{-} & \multicolumn{1}{l}{MSA} & \multicolumn{1}{l}{EA} \\ \midrule
\multicolumn{5}{l}{POS} \\ \midrule
HIN-ENG & 12,589 & 9,882 & HIN & ENG \\
SPA-ENG & 178,135 & 92,517 & SPA & ENG \\ \bottomrule
\end{tabular}
}
\caption{Dataset statistics are taken from~\citet{aguilar2020lince}. We define L1 and L2 as the languages found in the dataset. For example, in HIN-ENG, L1 is HIN and L2 is ENG. $^{\dagger}$We define MSA as ML and EA as EL. \#L1 represents the number of tokens in the first language and \#L2 represents the number of tokens in the second language.}
\label{dataset}
\end{table}

\subsection{Experimental Setup}
We describe our experimental details for each model.

\subsubsection{Scratch}
We train transformer-based models without any pre-training by following the mBERT model structure, and the parameters are randomly initialized, including the subword embeddings. We train transformer models with four and six layers with a hidden size of 768. This setting is important to measure the effectiveness of pre-trained multilingual models. We start the training with a learning rate of 1e-4 and an early stop of 10 epochs.

\subsubsection{Word Embeddings}
We use FastText embeddings~\cite{grave2018learning,mikolov2018advances} to train our transformer models. The model consists of a 4-layer transformer encoder with four heads and a hidden size of 200. 
We train a transformer followed by a Conditional Random Field (CRF) layer \cite{lafferty2001conditional}. The model is trained by starting with a learning rate of 0.1 with a batch size of 32 and an early stop of 10 epochs. We also train our model with only ML and EL embeddings. We freeze all embeddings and only keep the classifier trainable.

We leverage MUSE~\cite{lample2018word} to align the embeddings space between the ML and EL. MUSE mainly consists of two stages: adversarial training and a refinement procedure. For all alignment settings, we conduct the adversarial training using the SGD optimizer with a starting learning rate of 0.1, and then we perform the refinement procedure for five iterations using the Procrustes solution and CSLS~\cite{lample2018word}. After the alignment, we train our model with the aligned word embeddings (MUSE (ML $\rightarrow$ EL) or MUSE (EL $\rightarrow$ ML)) on the code-switching tasks.

\subsubsection{Pre-trained Multilingual Models}
We use pre-trained models from Huggingface.~\footnote{\href{https://github.com/huggingface/transformers}{https://github.com/huggingface/transformers}} On top of each model, we put a fully-connected layer classifier. We train the model with a learning rate between [1e-5, 5e-5] with a decay of 0.1 and a batch size of 8. For large models, such as $\text{XLM-R}_{\text{LARGE}}$ and $\text{XLM-MLM}_{\text{LARGE}}$, we freeze the embeddings layer to fit in a single GPU. 

\subsubsection{Multilingual Meta-Embeddings (MME)}
We use pre-trained word embeddings to train our MME. Table~\ref{embedding-list} shows the embeddings used for each dataset. We freeze all embeddings and train a transformer classifier with the CRF. The transformer classifier consists of a hidden size of 200, a head of 4, and 4 layers. All models are trained with a learning rate of 0.1, an early stop of 10 epochs, and a batch size of 32. We follow the implementation from the code repository.~\footnote{\href{https://github.com/gentaiscool/meta-emb}{https://github.com/gentaiscool/meta-emb}} Table~\ref{embedding-list} shows the list of word embeddings used in MME.
\begin{table}[!ht]
\centering
\resizebox{0.49\textwidth}{!}{
\begin{tabular}{ll}
\toprule
 & \textbf{Word Embeddings List} \\ \midrule
\multicolumn{2}{l}{NER} \\ \midrule
HIN-ENG & FastText: Hindi, English ~\cite{grave2018learning}\\
SPA-ENG & FastText: Spanish, English, Catalan,
\\
& Portugese~\cite{grave2018learning}\\ 
& GLoVe: English-Twitter~\cite{pennington2014glove} \\
MSA-EA & FastText: Arabic, Egyptian~\cite{grave2018learning} \\ \midrule
\multicolumn{2}{l}{POS} \\ \midrule
HIN-ENG & FastText: Hindi, English~\cite{grave2018learning} \\
SPA-ENG & FastText: Spanish, English, Catalan, \\
& Portugese~\cite{grave2018learning}\\ 
& GLoVe: English-Twitter~\cite{pennington2014glove} \\ \bottomrule
\end{tabular}
}
\caption{Embeddings list for MME.}
\label{embedding-list}
\end{table}

\subsubsection{Hierarchical Meta-Embeddings (HME)}
We train our HME model using the same embeddings as MME and pre-trained subword embeddings from~\citet{heinzerling2018bpemb}. The subword embeddings for each language pair are shown in Table~\ref{subword-embedding-list}.
\begin{table}[!ht]
\centering
\resizebox{0.45\textwidth}{!}{
\begin{tabular}{ll}
\toprule
 & \textbf{Subword Embeddings List} \\ \midrule
\multicolumn{2}{l}{NER} \\ \midrule
HIN-ENG & Hindi, English \\
SPA-ENG & Spanish, English, Catalan, Portugese\\ 
MSA-EA & Arabic, Egyptian \\ \midrule
\multicolumn{2}{l}{POS} \\ \midrule
HIN-ENG & Hindi, English \\
SPA-ENG & Spanish, English, Catalan, Portugese \\ 
\bottomrule
\end{tabular}
}
\caption{Subword embeddings list for HME.}
\label{subword-embedding-list}
\end{table}
We freeze all word embeddings and subword embeddings, and keep the character embeddings trainable.

\subsection{Other Baselines}
We compare the results with Char2subword and mBERT (cased) from~\citet{aguilar2020char2subword}. We also include the results of English BERT provided by the organizer of the LinCE public benchmark leaderboard (accessed on March 12nd, 2021).~\footnote{ \href{https://ritual.uh.edu/lince}{https://ritual.uh.edu/lince}}

\begin{table*}[!ht]
\centering
\resizebox{0.98\textwidth}{!}{
\begin{tabular}{lc|rlrlrl|rlrl}
\toprule
& \multicolumn{1}{c}{} & \multicolumn{6}{c}{NER} & \multicolumn{4}{c}{POS} \\
\cmidrule(l{2pt}r{2pt}){3-8} \cmidrule(l{2pt}r{2pt}){9-12}
& \multicolumn{1}{c}{} & \multicolumn{2}{c}{HIN-ENG} & \multicolumn{2}{c}{SPA-ENG} & \multicolumn{2}{c}{MSA-EA} & \multicolumn{2}{c}{HIN-ENG} & \multicolumn{2}{c}{SPA-ENG} \\
 \cmidrule(l{2pt}r{2pt}){3-4} \cmidrule(l{2pt}r{2pt}){5-6}
 \cmidrule(l{2pt}r{2pt}){7-8}
 \cmidrule(l{2pt}r{2pt}){9-10}
 \cmidrule(l{2pt}r{2pt}){11-12}
\multicolumn{1}{l}{\textbf{Method}} & \multicolumn{1}{l}{\textbf{Avg Perf.}} &\multicolumn{1}{|c}{Params} & \multicolumn{1}{c}{F1} &  \multicolumn{1}{c}{Params} & \multicolumn{1}{c}{F1} & \multicolumn{1}{c}{Params} & \multicolumn{1}{c|}{F1} & \multicolumn{1}{c}{Params} & \multicolumn{1}{c}{Acc} & \multicolumn{1}{c}{Params} & \multicolumn{1}{c}{Acc} \\
 \midrule
Scratch (2L) & 63.40 & 96M & 46.51 & 96M & 32.75 & 96M & 60.14 & 96M & 83.20 & 96M & 94.39 \\
Scratch (4L) & 60.93 & 111M & 47.01 & 111M & 19.06 & 111M & 60.24 & 111M & 83.72 & 111M & 94.64 \\ \midrule
\multicolumn{11}{l}{Mono/Multilingual Word Embeddings} \\ \midrule
FastText (ML) & 76.43 & 4M & 63.58 & 18M & 57.10 & 16M & 78.42 & 4M & 84.63 & 6M & 98.41 \\ 
FastText (EL) & 76.71 & 4M & 69.79 & 18M & 58.34 & 16M & 72.68 & 4M & 84.40 & 6M & 98.36 \\ 
MUSE (ML $\rightarrow$ EL) & 76.54 & 4M & 64.05 & 18M & 58.00 & 16M & 78.50 & 4M & 83.82& 6M & 98.34 \\
MUSE (EL $\rightarrow$ ML) & 75.58 & 4M & 64.86 & 18M & 57.08 & 16M & 73.95 & 4M & 83.62 & 6M & 98.38 \\\midrule
\multicolumn{11}{l}{Pre-Trained Multilingual Models} \\\midrule
mBERT (uncased) & 79.46 & 167M & 68.08 & 167M & 63.73 & 167M & 78.61 & 167M & 90.42 & 167M & 96.48 \\
mBERT (cased){$^\ddagger$} & 79.97 & 177M & 72.94 & 177M & 62.66 & 177M & 78.93 & 177M & 87.86 & 177M & 97.29 \\
Char2Subword{$^\ddagger$} & 81.07 & 136M & 74.91 & 136M & 63.32 & 136M & 80.45 & 136M & 89.64 & 136M & 97.03 \\
$\text{XLM-R}_{\text{BASE}}$ & 81.90 & 278M & 76.85 & 278M & 62.76 & 278M & 81.24 & 278M & 91.51 & 278M & 97.12 \\
XLM-R$_{\text{LARGE}}$ & \textbf{84.39} & 565M & \textbf{79.62} & 565M & \textbf{67.18} & 565M & \textbf{85.19} & 565M & 92.78 & 565M & 97.20 \\ 
$\text{XLM-MLM}_{\text{LARGE}}$ & 81.41 & 572M & 73.91 & 572M & 62.89 & 572M & 82.72 & 572M & 90.33 & 572M & 97.19 \\ \midrule
\multicolumn{11}{l}{Multilingual Meta-Embeddings} \\ \midrule
Concat & 79.70 & 10M & 70.76 & 86M & 61.65 & 31M & 79.33 & 8M & 88.14 & 23M & 98.61 \\
Linear & 79.60 & 10M & 69.68 & 86M & 61.74 & 31M & 79.42 & 8M & 88.58 & 23M & 98.58 \\
Attention (MME) & 79.86 & 10M & 71.69 & 86M & 61.23 & 31M & 79.41 & 8M & 88.34 & 23M & 98.65 \\
HME & 81.60 & 12M & 73.98 & 92M & 62.09 & 35M & 81.26 & 12M & 92.01 & 30M & 98.66 \\
HME-Ensemble & \textbf{82.44} & 20M & 76.16 & 103M & 62.80 & 43M & 81.67 & 20M &	\textbf{92.84} & 40M & \textbf{98.74} \\ \bottomrule
% \midrule
% \multicolumn{11}{l}{Mixture-of-Language Models} \\
% $\text{mBERT} + \text{XLM-R}_{\text{BASE}}$ (Attention) & & & & & & & & & & \\
% $\text{mBERT} + \text{XLM-R}_{\text{BASE}}$ (Concat) & & & & & & & & & & \\
% $\text{mBERT} + \text{XLM-R}_{\text{BASE}}$ (Linear) & & & & & & & & & & \\\bottomrule
\end{tabular}
}
\caption{Results on the development set of the LinCE benchmark. $^{\ddagger}$ The results are taken from~\citet{aguilar2020char2subword}. The number of parameters of mBERT (cased) is calculated by approximation.}
\label{table:dev_result}
\end{table*}

\begin{table*}[!ht]
\centering
\resizebox{0.98\textwidth}{!}{
\begin{tabular}{lrc|ccc|cc}
\toprule
& \multicolumn{2}{c}{} & \multicolumn{3}{c}{NER} & \multicolumn{2}{c}{POS} \\
\cmidrule(l{2pt}r{2pt}){4-6} \cmidrule(l{2pt}r{2pt}){7-8}
\multicolumn{1}{l}{\textbf{Method}} & \multicolumn{1}{c}{\textbf{Avg Params}} & \multicolumn{1}{l}{\textbf{Avg Perf.$\uparrow$}} &  \multicolumn{1}{c}{HIN-ENG} & \multicolumn{1}{c}{SPA-ENG} & \multicolumn{1}{c}{MSA-EA} & \multicolumn{1}{c}{HIN-ENG} & \multicolumn{1}{c}{SPA-ENG} \\ \midrule
English BERT (cased)$^{\dagger}$ & 108M & 75.80 & 74.46 & 61.15 & 59.44 & 87.02 & 96.92 \\
mBERT (cased)$^{\ddagger}$ & 177M & 77.08 & 72.57 & 64.05 & 65.39 &  86.30 & 97.07 \\
HME & 36M & 77.64 & 73.78 & 63.06 & 66.14 & 88.55 & 96.66 \\
Char2Subword$^{\ddagger}$ & 136M & 77.85 & 73.38 & 64.65 & 66.13 & 88.23 & 96.88 \\
XLM-MLM$_{\text{LARGE}}$ & 572M & 78.40 & 74.49 & 64.16 & 67.22 & 89.10 & 97.04 \\
XLM-R$_{\text{BASE}}$ & 278M & 78.75 & 75.72 & 64.95 & 65.13 & 91.00 & 96.96 \\
HME-Ensemble & \underline{45M} & \underline{79.17} & 75.97 & 65.11 & \textbf{68.71} & 89.30 & 96.78 \\
XLM-R$_{\text{LARGE}}$ & 565M & \textbf{80.96} & \textbf{80.70} & \textbf{69.55} & 65.78 & \textbf{91.59} & \textbf{97.18} \\ 
\bottomrule
% \midrule
% \multicolumn{11}{l}{Mixture-of-Language Models} \\
% $\text{mBERT} + \text{XLM-R}_{\text{BASE}}$ (Attention) & & & & & & & & & & \\
% $\text{mBERT} + \text{XLM-R}_{\text{BASE}}$ (Concat) & & & & & & & & & & \\
% $\text{mBERT} + \text{XLM-R}_{\text{BASE}}$ (Linear) & & & & & & & & & & \\\bottomrule
\end{tabular}
}
\caption{Results on the test set of the LinCE benchmark.$^{\ddagger}$ The results are taken from~\citet{aguilar2020char2subword}. $^{\dagger}$ The result is taken from the LinCE leaderboard.}
\label{table:test_result}
\end{table*}

\section{Results and Discussions}

\subsection{LinCE Benchmark}
We evaluate all the models on the LinCE benchmark, and the development set results are shown in Table \ref{table:dev_result}. As expected, models without any pre-training (e.g., Scratch (4L)) perform significantly worse than other pre-trained models. Both FastText and MME use pre-trained word embeddings, but MME achieves a consistently higher F1 score than FastText in both NER and POS tasks, demonstrating the importance of the contextualized self-attentive encoder. HME further improves on the F1 score of the MME models, suggesting that encoding hierarchical information from sub-word level, word level, and sentence level representations can improve code-switching task performance. Comparing HME with contextualized pre-trained multilingual models such as mBERT and XLM-R, we find that HME models are able to obtain competitive F1 scores while maintaining a 10x smaller model sizes. This result indicates that pre-trained multilingual word embeddings can achieve a good balance between performance and model size in code-switching tasks. Table \ref{table:test_result} shows the models' performance in the LinCE test set. The results are highly correlated to the results of the development set. $\text{XLM-R}_{\text{LARGE}}$ achieves the best-averaged performance, with a 13x larger model size compared to the HME-Ensemble model.

\subsection{Model Effectiveness and Efficiency}
\begin{figure*}
    \centering
    \resizebox{0.47\textwidth}{!}{
        \includegraphics{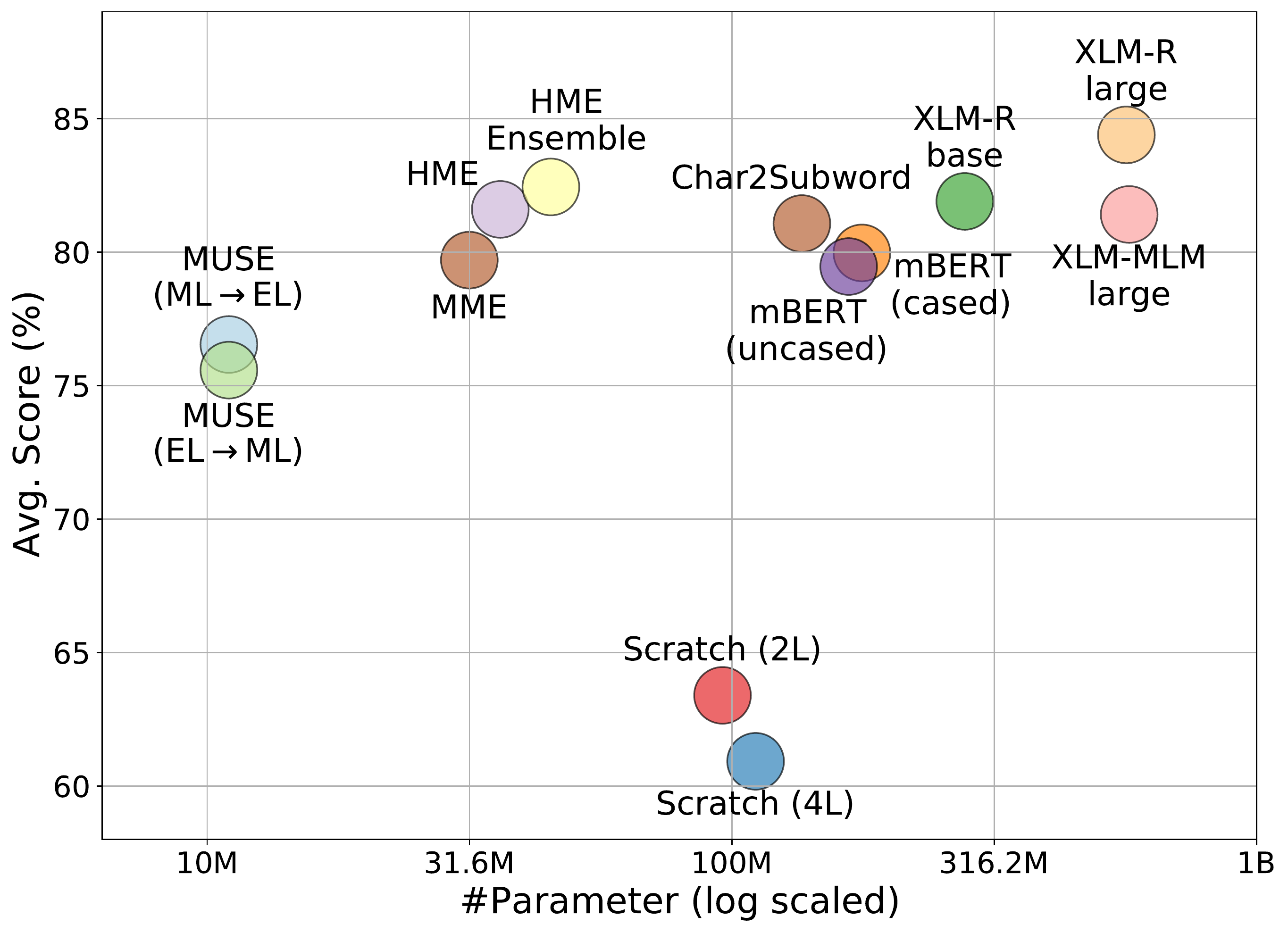}
    }
    \hspace{10pt}
    \resizebox{0.47\textwidth}{!}{
        \includegraphics{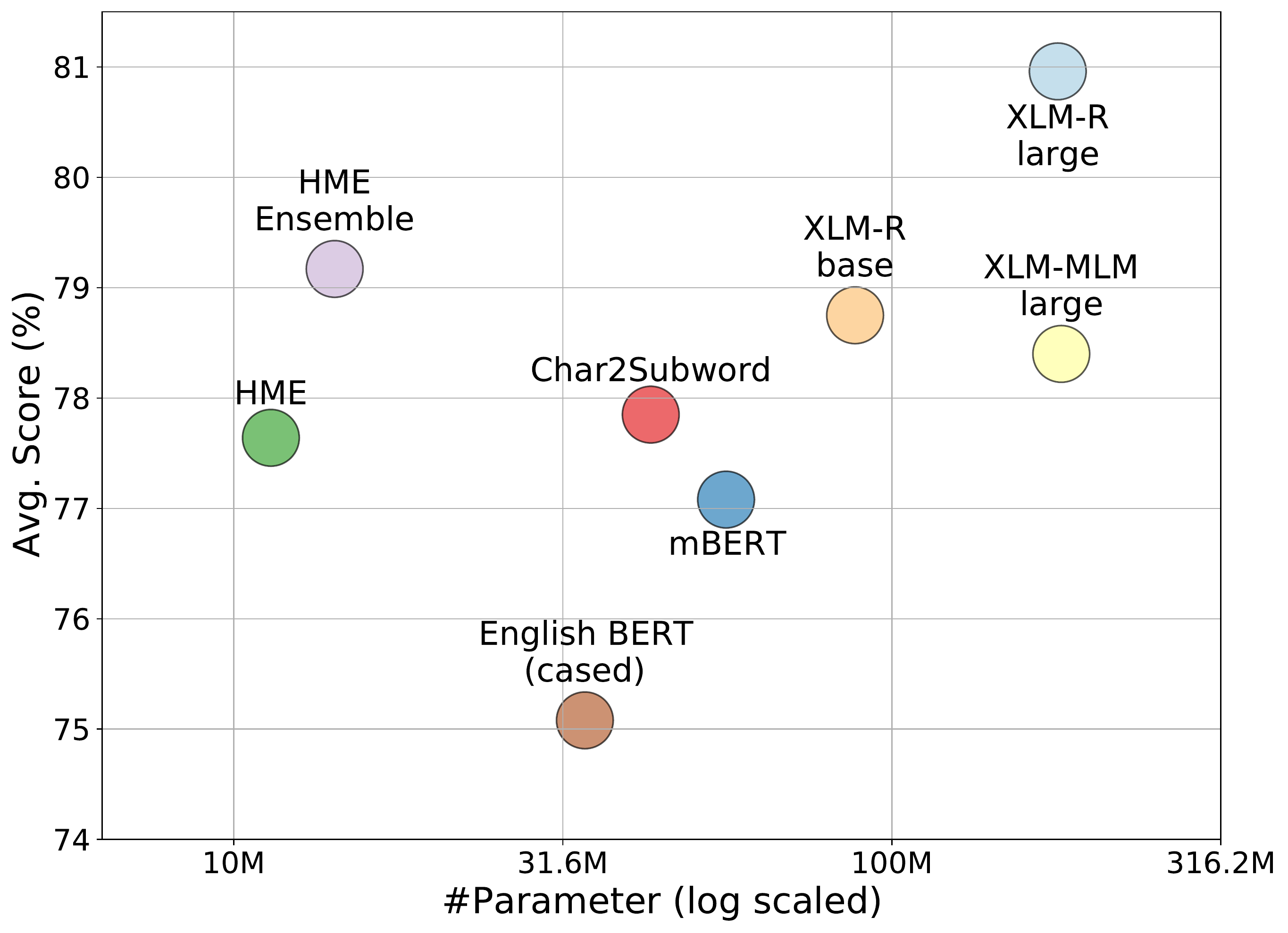}
    }
    \caption{Validation set (left) and test set (right) evaluation performance (y-axis) and parameter (x-axis) of different models on LinCE benchmark.}
    \label{fig:performance_vs_parameters}
\end{figure*}
\paragraph{Performance vs. Model Size} As shown in Figure \ref{fig:performance_vs_parameters}, the Scratch models yield the worst average score, at around 60.93 points. With the smallest pre-trained embedding model, FastText, the model performance can improve by around 10 points compared to the Scratch models and they only have 10M parameters on average. On the other hand, the MME models, which have 31.6M parameters on average, achieve similar results to the mBERT models, with around 170M parameters. Interestingly, adding subwords and character embeddings to MME, such as in the HME models, further improves the performance of the MME models and achieves a 81.60 average score, similar to that of the XLM-R$_{\text{BASE}}$ and XLM-MLM$_{\text{LARGE}}$ models, but with less than one-fifth the number of parameters, at around 42.25M. The Ensemble method adds further performance improvement of around 1\% with an additional 2.5M parameters compared to the non-Ensemble counterparts.

\paragraph{Inference Time} To compare the speed of different models, we use generated dummy data with various sequence lengths, [16, 32, 64, 128, 256, 512, 1024, 2048, 4096]. We measure each model's inference time and collect the statistics of each model at one particular sequence length by running the model 100 times. The experiment is performed on a single NVIDIA GTX1080Ti GPU. We do not include the pre-processing time in our analysis. Still, it is clear that the pre-processing time for meta-embeddings models is longer than for other models as pre-processing requires a tokenization step to be conducted for the input multiple times with different tokenizers. The sequence lengths are counted based on the input tokens of each model. We use words for the MME and HME models, and subwords for other models.
% As the inputs of FastText, MME, and HME models are in word-level, and other implementations are based on subwords (except for Char2Subword since the implementation code has not been released yet), we apply the sequence length for word-level in MME and HME models and subword-level in the other models. 
% As the inputs of MME and HME models, we use two BPE tokens and five character tokens.
% For each word in the dummy inputs of MME and HME models, we use two BPE tokens and five character tokens. This means that both MME and HME models encode double the other models' size on our analysis, which might be enough to compensate for the preprocessing time for the meta embedding model. As MME and HME models have different configurations for each task, we measure the inference time from all the configurations, each with 100 iterations, and take the combined statistics over all the configurations.

\begin{figure}
    \centering
    \resizebox{0.49\textwidth}{!}{
        \includegraphics{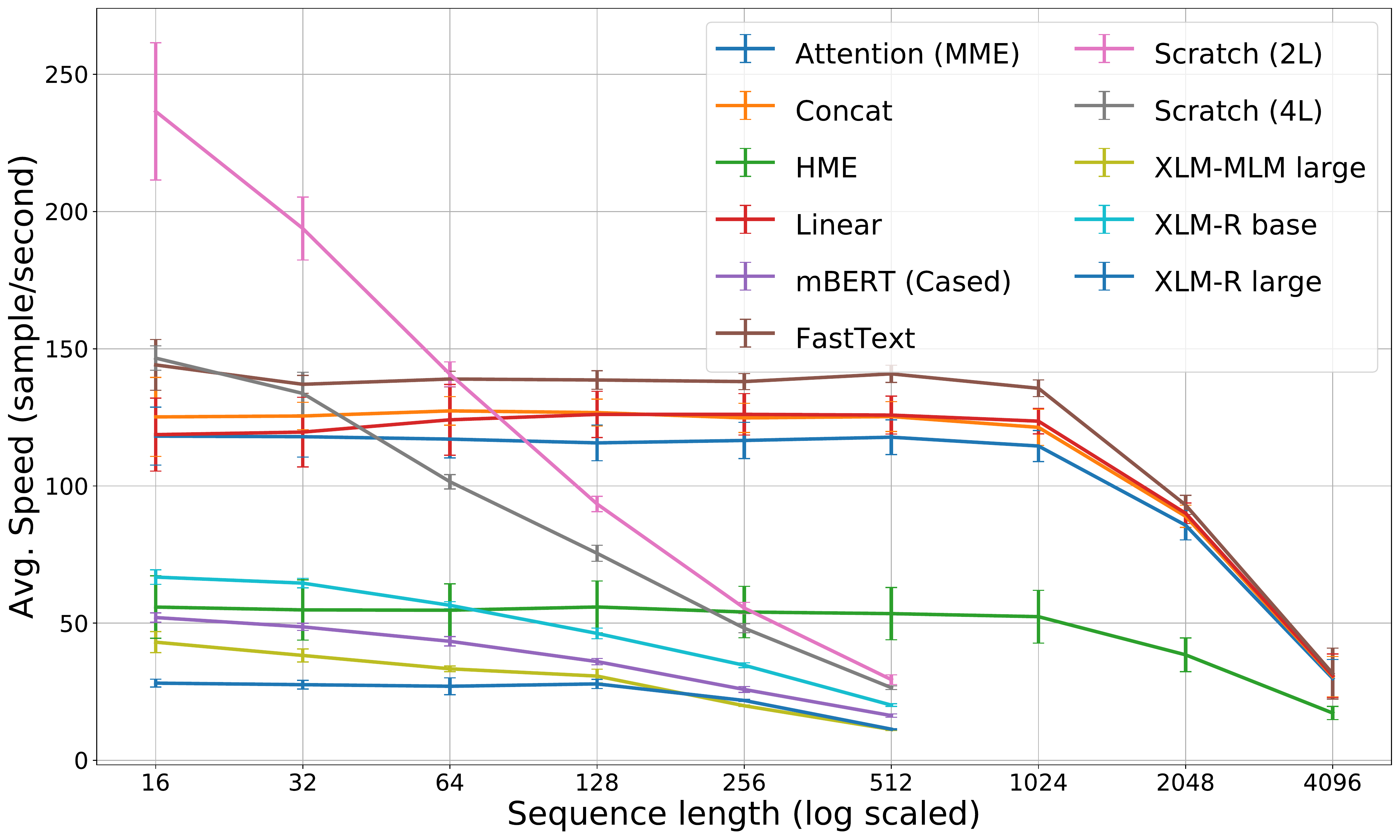}
    }
    \caption{Speed-to-sequence length comparison of different models.}
    \label{fig:speed_vs_seqlen}
\end{figure}

The results of the inference speed test are shown in Figure \ref{fig:speed_vs_seqlen}. Although all pre-trained contextualized language models yield a very high validation score, these models are also the slowest in terms of inference time. For shorter sequences, the HME model performs as fast as the mBERT and XLM-R$_{\text{BASE}}$ models, but it can retain the speed as the sequence length increases because of the smaller model dimension in every layer. The FastText, MME, and Scratch models yield a high throughput in short-sequence settings by processing more than 150 samples per second. For longer sequences, the same behavior occurs, with the throughput of the Scratch models reducing as the sequence length increases, even becoming lower than that of the HME model when the sequence length is greater than or equal to 256. Interestingly, for the FastText, MME, and HME models, the throughput remains steady when the sequence length is less than 1024, and it starts to decrease afterwards.

\paragraph{Memory Footprint} We record the memory footprint over different sequence lengths, and use the same setting for the FastText, MME, and HME models as in the inference time analysis. We record the size of each model on the GPU and the size of the activation after performing one forward operation to a single sample with a certain sequence length. The result of the memory footprint analysis for a sequence length of 512 is shown in Table \ref{tab:memory_512}. Based on the results, we can see that meta-embedding models use a significantly smaller memory footprint to store the model and activation memory. For instance, the memory footprint of the HME model is less than that of the Scratch (4L) model, which has only four transformer encoder layers, a model dimension of 768 and a feed-forward dimension of 3,072. On the other hand, large pre-trained language models, such as XLM-MLM$_{\text{LARGE}}$ and XLM-R$_{\text{LARGE}}$, use a much larger memory for storing the activation memory compared to all other models. The complete results of the memory footprint analysis are shown in Appendix A.

\begin{table}[ht!]
\centering
\resizebox{0.4\textwidth}{!}{
    \begin{tabular}{lrr}
    \toprule
    \textbf{Model} &  \textbf{Activation (MB)} \\
    \midrule
    FastText &               79.0 \\
    Concat &               85.3 \\
    Linear &                80.8 \\
    Attention (MME) &      88.0 \\
    HME &           154.8 \\
    Scratch (2L) &             133.0 \\
    Scratch (4L) &            264.0 \\
    mBERT &            597.0 \\
    XLM-R$_{\text{BASE}}$ &           597.0 \\
    XLM-R$_{\text{LARGE}}$ &           1541.0 \\
    XLM-MLM$_{\text{LARGE}}$ &        1158.0 \\
    \bottomrule
    \end{tabular}
}
\caption{GPU memory consumption of different models with input size of 512.}
\label{tab:memory_512}
\end{table}

% \subsection{Training Data Size on Pre-trained Multilingual Models}
% % Fast-Text lebih kecil vs lebih besar

% \subsection{Transliteration on Hindi}

\section{Related Work}

\paragraph{Transfer Learning on Code-Switching}
Previous works on code-switching have mostly focused on combining pre-trained word embeddings with trainable character embeddings to represent noisy mixed-language text~\cite{trivedi2018iit,wang2018code,winata2018bilingual}.~\citet{winata2018code} presented a multi-task training framework to leverage part-of-speech information in a language model. Later, they introduced the MME in the code-switching domain by combining multiple word embeddings from different languages~\cite{winata2019learning}. MME has since also been applied to Indian languages~\cite{priyadharshini2020named,dowlagar2021cmsaone}.

Meta-embeddings have been previously explored in various monolingual NLP tasks~\cite{yin2016learning,muromagi2017linear,bollegala2018think,coates2018frustratingly,kiela2018dynamic}.~\citet{winata2019hierarchical} introduced hierarchical meta-embeddings by leveraging subwords and characters to improve the code-switching text representation. \citet{pratapa-etal-2018-word} propose to train skip-gram embeddings from synthetic code-switched data generated by~\citet{pratapa2018language}. This improves syntactic and semantic code-switching tasks.~\citet{winata2018learn,lee2019linguistically,winata2019code,samanta2019deep}, and~\citet{gupta2020semi} proposed a generative-based model for augmenting code-switching data from parallel data. Recently, ~\citet{aguilar2020char2subword} proposed the Char2Subword model, which builds representations from characters out of the subword vocabulary, and they used the module to replace subword embeddings that are robust to misspellings and inflection that are mainly found in a social media text.~\citet{khanuja2020gluecos} explored fine-tuning techniques to improve mBERT for code-switching tasks, while~\citet{winata2020meta} introduced a meta-learning-based model to leverage monolingual data effectively in code-switching speech and language models. 

\paragraph{Bilingual Embeddings}
In another line of works, bilingual embeddings have been introduced to represent code-switching sentences, such as in bilingual correlation-based embeddings (BiCCA)~\cite{faruqui2014improving}, the bilingual compositional model (BiCVM)~\cite{hermann-blunsom-2014-multilingual}, BiSkip~\cite{luong-etal-2015-bilingual}, RCSLS~\cite{joulin2018loss}, and MUSE~\cite{lample2017unsupervised,lample2018word}, to align words in L1 to the corresponding words in L2, and vice versa. 

\section{Conclusion}
In this paper, we study multilingual language models' effectiveness so as to understand their capability and adaptability to the mixed-language setting. We conduct experiments on named entity recognition and part-of-speech tagging on various language pairs. We find that a pre-trained multilingual model does not necessarily guarantee high-quality representations on code-switching, while the hierarchical meta-embeddings (HME) model achieve similar results to mBERT and $\text{XLM-R}_{\text{BASE}}$ but with significantly fewer parameters. Interestingly, we find that $\text{XLM-R}_{\text{LARGE}}$ has better performance by a great margin, but with a substantial cost in the training and inference time, using 13x more parameters than HME-Ensemble for only a 2\% improvement. 

\section*{Acknowledgments}
This work has been partially funded by ITF/319/16FP and MRP/055/18 of the Innovation Technology Commission, the Hong Kong SAR Government, and School of Engineering Ph.D. Fellowship Award, the Hong Kong University of Science and Technology, and RDC 1718050-0 of EMOS.AI.

% Entries for the entire Anthology, followed by custom entries
\bibliography{anthology,custom}
\bibliographystyle{acl_natbib}

\newpage

\appendix

\section{Memory Footprint Analysis}
\label{sec:appendix}
We show the complete results of our memory footprint analysis in Table \ref{tab:memory_all}.
\begin{table*}[!ht]
\centering
\resizebox{0.98\textwidth}{!}{
\begin{tabular}{l|rrrrrrrrr}
\toprule
\multirow{2}{*}{\textbf{Model}} &  \multicolumn{9}{c}{\textbf{Activation (MB)}} \\
 & \textbf{16}   &    \textbf{32}  & \textbf{64}  & \textbf{128} & \textbf{256} & \textbf{512} & \textbf{1024} & \textbf{2048} & \textbf{4096} \\
\midrule
FastText   &   1.0 &   2.0 &    4.0 &   10.0 &   26.0 &    79.0 &  261.0 &   941.0 &  3547.0 \\
Linear  &   1.0 &   2.0 &    4.0 &   10.0 &   27.4 &    80.8 &  265.6 &   950.0 &  3562.0 \\
Concat   &   1.0 &   2.0 &    5.0 &   11.2 &   29.2 &    85.2 &  274.5 &   967.5 &  3596.5 \\
Attention (MME)  &   1.0 &   2.0 &    5.4 &   12.4 &   31.0 &    89.0 &  283.2 &   985.6 &  3630.6 \\
HME  &   3.2 &   6.6 &   13.4 &   28.6 &   64.2 &   154.8 &  416.4 &  1252.0 &  4155.0 \\
Scratch (2L) &   2.0 &   4.0 &    8.0 &   20.0 &   46.0 &   133.0 &         - &          - &          - \\
Scratch (4L)  &   3.0 &   7.0 &   15.0 &   38.0 &   90.0 &   264.0 &         - &          - &          - \\
mBERT (uncased) &  10.0 &  20.0 &   41.0 &  100.0 &  218.0 &   597.0 &         - &          - &          - \\
XLM-R$_{\text{BASE}}$ &  10.0 &  20.0 &   41.0 &  100.0 &  218.0 &   597.0 &         - &          - &          - \\
XLM-R$_{\text{LARGE}}$ &  25.0 &  52.0 &  109.0 &  241.0 &  579.0 &  1541.0 &         - &          - &          - \\
XLM-MLM$_{\text{LARGE}}$ &  20.0 &  42.0 &   89.0 &  193.0 &  467.0 &  1158.0 &         - &          - &          - \\
\bottomrule
\end{tabular}
}
\caption{Memory footprint (MB) for storing the activations for a given sequence length.}
\label{tab:memory_all}
\end{table*}

\end{document}